\pdfoutput=1
\documentclass[11pt]{article}

\usepackage[preprint]{coling}

\usepackage{times}
\usepackage{latexsym}

\usepackage[T1]{fontenc}
\usepackage[utf8]{inputenc}

\usepackage{microtype}

\usepackage{inconsolata}

\usepackage{graphicx}
\newcommand*\samethanks[1][\value{footnote}]{\footnotemark[#1]}
\usepackage{multirow}
\usepackage{multicol}
\usepackage{float}
\usepackage{amsmath} 
\usepackage{colortbl}
\usepackage{booktabs}
\usepackage{enumitem}

\title{CJEval: A Benchmark for Assessing Large Language Models Using Chinese Junior High School Exam Data}

\author{Qian-Wen Zhang$^1$\thanks{Equal contribution.}, Haochen Wang$^2$\samethanks\thanks{Work done during an internship at Tencent.}, Fang Li$^1$, Siyu An$^1$, Lingfeng Qiao$^1$,\\
\textbf{Liangcai Gao}$^2$, \textbf{Di Yin}$^1$ and \textbf{Xing Sun}$^1$ \\
        $^1$Tencent YouTu Lab, $^2$Peking University\\ 
        $^1$\{cowenzhang, frankfangli, siyuan, leafqiao, endymecyyin, winfredsun\}@tencent.com \\
        $^2$2201213129@stu.pku.edu.cn, gaoliangcai@pku.edu.cn
        }

\begin{document}
\maketitle
\begin{abstract}
Online education platforms have significantly transformed the dissemination of educational resources by providing a dynamic and digital infrastructure. With the further enhancement of this transformation, the advent of Large Language Models (LLMs) has elevated the intelligence levels of these platforms. However, current academic benchmarks provide limited guidance for real-world industry scenarios. This limitation arises because educational applications require more than mere test question responses. To bridge this gap, we introduce \textbf{CJEval}\footnote{Available at \url{https://github.com/SmileWHC/CJEval}.}, a benchmark based on \textbf{C}hinese \textbf{J}unior High School Exam \textbf{Eval}uations. CJEval consists of 26,136 samples across four application-level educational tasks covering ten subjects. These samples include not only questions and answers but also detailed annotations such as question types, difficulty levels, knowledge concepts, and answer explanations. By utilizing this benchmark, we assessed LLMs' potential applications and conducted a comprehensive analysis of their performance by fine-tuning on various educational tasks. Extensive experiments and discussions have highlighted the opportunities and challenges of applying LLMs in the field of education.

% \footnote{The dataset and relevant code are available at \url{https://github.com/xxxx/CJEval.}}

% \footnote{The dataset and relevant code are available.}

\end{abstract}

\section{Introduction}
Traditional educational paradigms, which predominantly rely on human instructors, often face limitations in scalability and resource allocation. The rise of online education platforms has mitigated some of these constraints by democratizing access to knowledge and skills through digital means. The emergence of Large Language Models (LLMs), such as ChatGPT \citep{brown2020language,achiam2023gpt} and Llama \citep{touvron2023llama} , signifies a transformative leap in the field of artificial intelligence, demonstrating an outstanding mastery of human language. In the context of education, these models are anticipated to fundamentally reshape the technological foundations of teaching and learning on digital platforms. Despite the promising prospects, implementing LLM-based educational systems presents significant challenges \citep{kasneci2023chatgpt,li2023adapting}. These models must be adept at not only understanding the specific issues encountered by students but also at applying specialized pedagogical knowledge to offer effective solutions. This necessitates a sophisticated interplay between linguistic comprehension and domain-specific expertise, ensuring that educational interventions are both accurate and pedagogically sound.

\begin{figure}[t] 
  \includegraphics[width=\columnwidth]{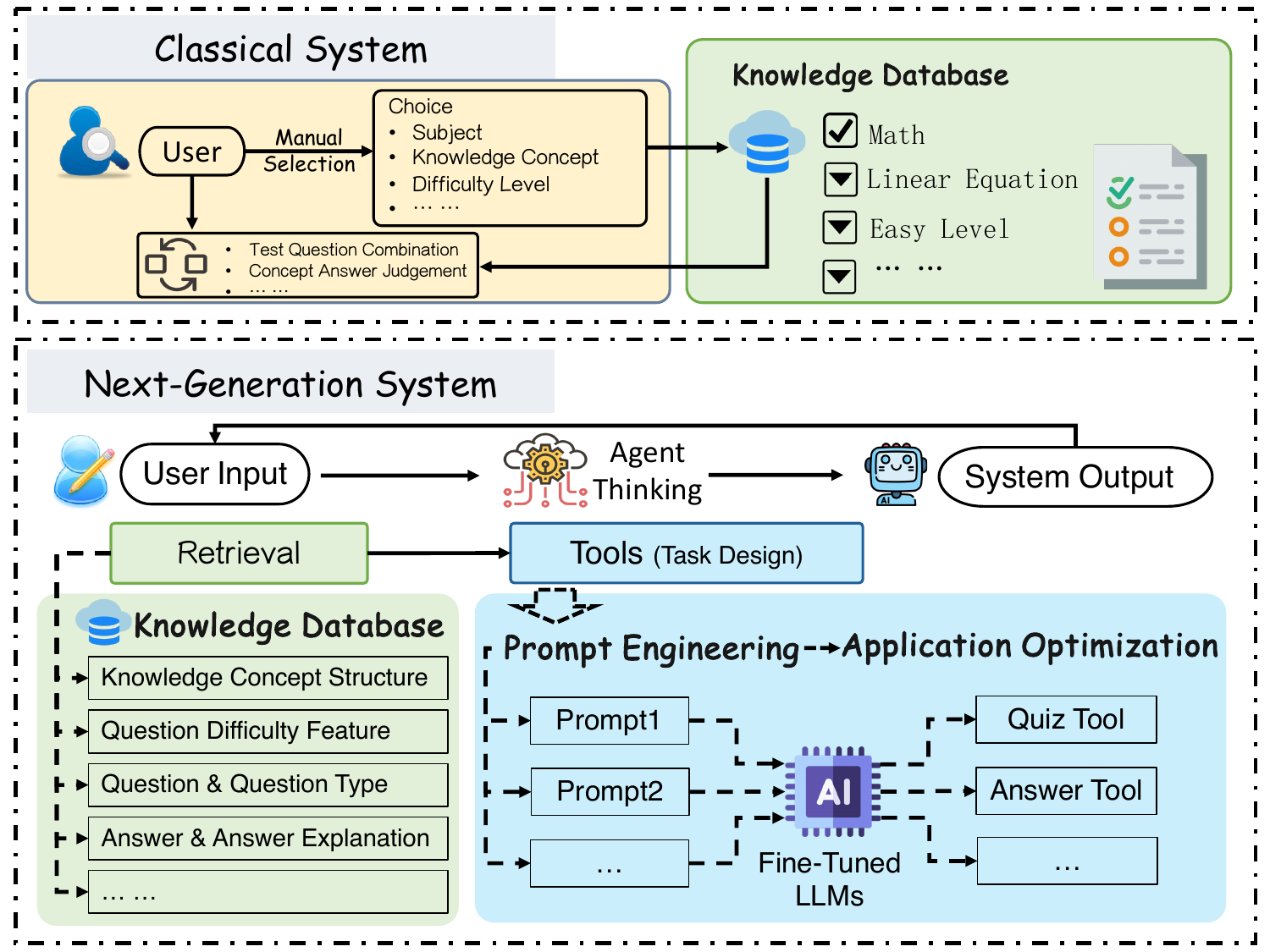}
  \caption{An overview of the design of online education systems reveals that next-generation systems are significantly more intelligent than classical ones.}
  \label{fig:system}
\end{figure}

Exam questions are the most critical data in educational settings, serving as the primary means to evaluate and measure students' understanding of knowledge \citep{glaser2001knowing}. Figure~\ref{fig:system} illustrates the differences between classical and next-generation education systems. Regarding the next-generation system, we examine LLMs from the perspective of utilizing exam resources to enable usable tools and explore the potential of LLM-based education systems, aiming to provide valuable insights into areas where further improvements are necessary. It should be noted that existing datasets for LLMs primarily focus on the accuracy of models in answering exam questions. For instance, benchmark tests such as MMLU \citep{hendrycks2020measuring} and M3Exam \citep{zhang2024m3exam}, as well as Chinese benchmarks like CMMLU \citep{li2023cmmlu} and M3KE \citep{liu2023m3ke}, are commonly employed. However, these benchmarks predominantly focus on a single type of question: multiple-choice questions. While facilitating the automated evaluation of LLMs, this focus does not fully reflect the models' comprehensive capabilities in educational assessments. This is because multiple-choice questions typically require only simple judgments, which may lead models to take shortcuts in decision-making. 
In contrast, early tasks, such as knowledge concept tagging and question difficulty prediction \citep{chen2014tag,qiu2019question}, often utilize datasets that focus solely on their specific task labels, lacking comprehensive multi-dimensional annotation information.

By integrating diverse tasks and data annotations, we aim to create a more comprehensive and robust assessment reference dataset for educational LLMs. To this end, a novel benchmark named \textbf{CJEval} was developed. CJEval, based on real Chinese Junior High School exam questions, includes not only questions and answers but also detailed information about question types, difficulty levels, knowledge concepts, and answer explanations. It simultaneously provides four core tasks: knowledge concept tagging, question difficulty prediction, question answering, and question generation.

In summary, the main contributions of this paper are as follows:
\begin{itemize} 
    \item A new evaluation benchmark called \textbf{CJEval} has been developed, specifically curated from data on Chinese Junior High School Examinations. It features the most comprehensive annotation information centered on exam questions, encompassing four application-level tasks across ten subjects, thus offering a robust basis for assessing educational competencies.
    \item Extensive testing was conducted on a broad range of the latest LLMs, and a thorough analysis was performed through fine-tuning. This detailed examination helped in identifying the potential applications as well as the limitations of LLMs within the educational sector. 
\end{itemize}

\section{Online Examination System} 
In this section, we first introduce the online examination system and then formally describe the details of the four application-level tasks, noting that they are all question-based.
\subsection{System Design}
For the next-generation system, a comprehensive knowledge database centered around test questions is essential. As shown in Figure~\ref{fig:system}, the next-generation system provides a smarter system control center and more flexible tools than classic system designs that focus on engineering logic. Its versatility allows developers to build LLM applications through multiple agents that can communicate with each other to accomplish tasks \citep{wu2023autogen}.
The retrieval process, inspired by retrieval-augmented generation (RAG) \citep{lewis2020retrieval}, is optional. Whether or not retrieval results are utilized, this module supports agent selection and activates downstream tools. These downstream tools comprise prompt engineering and application optimization. Prompt engineering entails refining and structuring the input prompts to enhance system performance, while application optimization involves integrating fine-tuned LLMs to execute various tasks. This approach ensures the effectiveness and quality of the final output. The modular and sophisticated design of the system enables it to adapt to diverse educational needs, ensuring users receive precise outputs.

\subsection{Question-Based Tasks}
In general, each test question is semi-structured data that contains multiple data fields, including the question stem, answer, and some metadata. Formally, we define the set of question stems as $Q=\{q_1,\dots,q_n\}$, the set of answers as $A=\{a_1,\dots,a_n\}$, and the set of answer explanations as $AE=\{e_1,\dots,e_n\}$. Additionally, the difficulties of questions are categorized into five levels $Dif=\{1,2,3,4,5\}$, and the set of knowledge concepts is represented by $K=\{k_1,\dots,k_m\}$.

\subsubsection{Knowledge Concept Tagging}

% https://www.processon.com/diagraming/665bc6b2c4b7591bd2649800
\begin{figure*}[bthp!]
  \centering
  \includegraphics[width=0.98\textwidth]{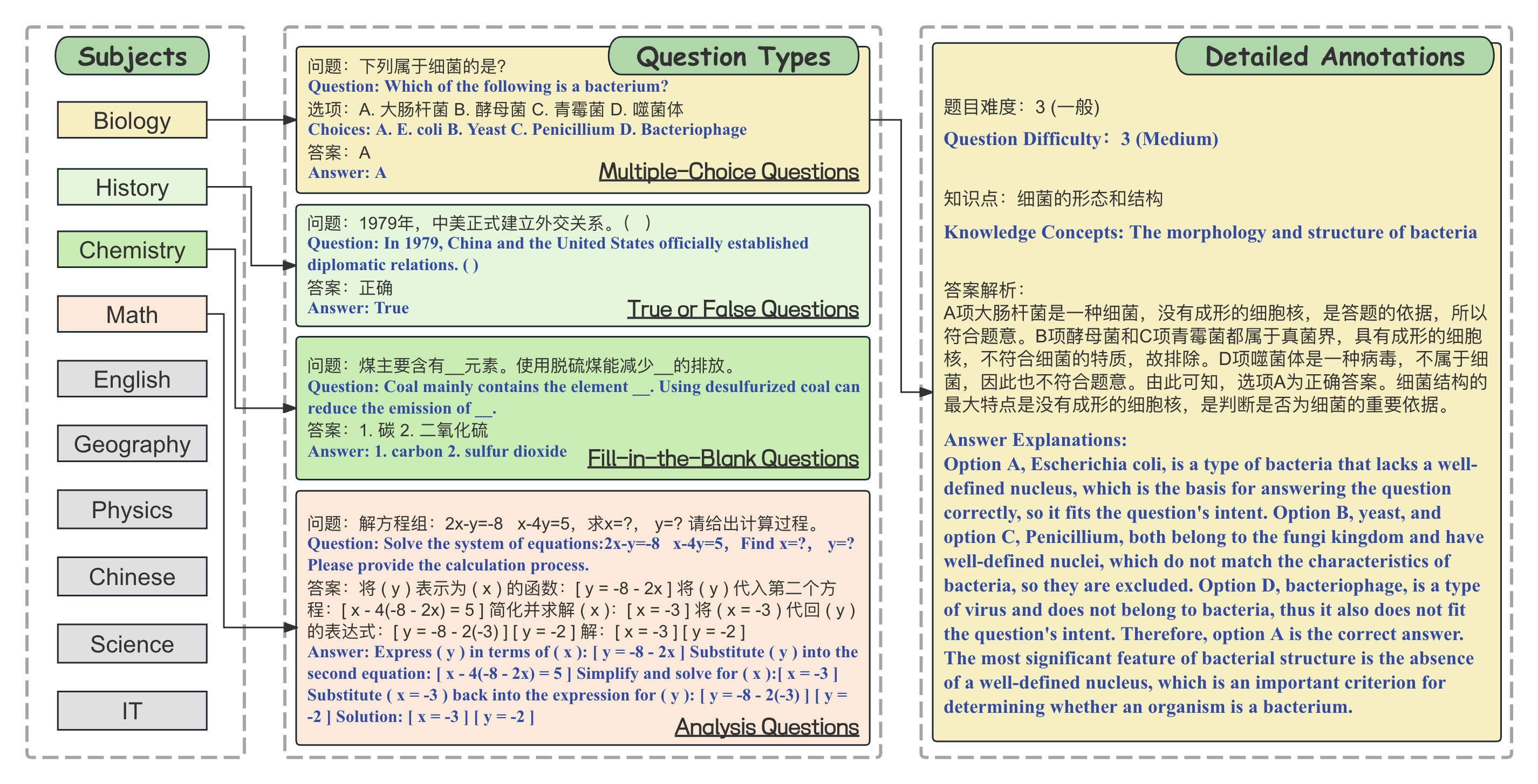}
  \caption{Examples in \textbf{CJEval}. CJEval comprises authentic junior high school exam questions across 10 subjects, featuring various question types, question difficulty levels, knowledge concepts, and answer explanations. English translations are shown below the corresponding Chinese texts for better readability.}
  \label{fig:showdata}
\end{figure*}

The task of knowledge concept tagging (KCT) \citep{chen2014tag,sun2018automatic,li2024automate} can be defined as follows: Given a particular knowledge concept $k_i$ from a set $K$ and a question $q_j$ from a set $Q$, the goal of the tagging model is to determine a binary outcome $y \in \{0, 1\}$. This outcome signifies whether $q_j$ corresponds to $k_i$. Previous studies have tackled this problem by transforming both the question and the knowledge concept into dense vector representations using various embedding techniques. 
In this paper, we default to reducing the candidate set of knowledge concepts by computing the similarity scores between questions and knowledge concepts.
Subsequently, using task-specific prompts these models generate answers that align with the most relevant knowledge concepts. This task is viewed as a multi-label classification problem.
% This two-step approach leverages the strengths of LLMs in understanding and contextualizing complex language inputs, leading to more accurate and efficient tagging outcomes. 

\subsubsection{Question Difficulty Prediction}
Question difficulty prediction (QDP) \citep{qiu2019question,alkhuzaey2023text,lee2023difficulty} aims to estimate the level of difficulty of a given question, often quantified as the percentage of students who fail to answer it correctly. The difficulty of a question is always related to the problem solver, so our labels are derived from historical student user error rate statistics.
Formally, question difficulties are categorized into five distinct levels: \{easy, relatively easy, medium, relatively difficult, difficult\}. This task is conceptualized as a multi-class classification problem.

\subsubsection{Question Answering}
The main objective of question answering (QA) \citep{hendrycks2020measuring,liu2024lhmke} is to provide an accurate answer $a_i$ to the given question $q_i$. The complexity of QA tasks varies significantly across different types of questions. For instance, multiple-choice questions typically require the model to select the correct option from a predefined set, which may sometimes lead to correct answers being chosen by chance. In contrast, analytical open-ended questions, like problem-solving, need detailed and structured responses, demonstrating knowledge, synthesis, and reasoning. 

\subsubsection{Question Generation}
Question generation (QG) \citep{artsi2024large} is a pivotal task in the domain of educational technology, particularly for the expansion and updating of question databases. The primary objective of QG is to create new, high-quality questions that can be used for assessments, practice, and learning reinforcement. Formally, given a question $q_i$, its corresponding answer $a_i$, the associated knowledge concepts $K_{q_i}$, and the difficulty level $dif_{i}$,  the goal of QG is to generate novel questions that are contextually relevant and pedagogically valuable. 

\section{CJEval Benchmark}
CJEval is collected and processed from authentic junior high school examination questions. This dataset encompasses a wide range of subjects and question types, ensuring comprehensive coverage of the curriculum. Each question is meticulously annotated with relevant metadata, including question type, difficulty level, associated knowledge concepts, and detailed answer explanations. 

Figure~\ref{fig:showdata} shows examples from CJEval. Specifically, CJEval features a diverse array of question types, categorized primarily into four groups. The first group, \textbf{Multiple-Choice Questions (MCQs)}, is a staple in educational assessments, designed to gauge a broad spectrum of knowledge and skills. Within this category, we distinguish between \textbf{Single-Choice Questions (SCQs)}, where respondents select one correct answer from multiple options, and \textbf{Multiple-Response Questions (MRQs)}, which permit the selection of more than one correct answer. The second category comprises \textbf{True or False Questions (TFQs)}, which assess the respondents' ability to discern the veracity of given statements. The third category includes \textbf{Fill-in-the-Blank Questions (FBQs)}, which require respondents to provide specific information or terms to complete a statement accurately. The fourth and final category, \textbf{Analysis Questions (AQs)}, encompasses Short Answer Questions and Calculation Questions, with the Chinese subject also featuring Modern Text Reading and Poetry Appreciation tasks.

\begin{table}
  \centering
  \small
  \setlength{\tabcolsep}{4pt}
  \begin{tabular}{lccccc}
\toprule[1.0pt]
    & \textbf{SCQs} & \textbf{MRQs} & \textbf{TFQs} & \textbf{FBQs} & \textbf{AQs} \\
    \hline
    No.S     & 10 & 5 & 5 & 9 & 7           \\
    No.Q     & 7,701 & 2,569 & 3,729 & 6,193 & 5,944          \\
    Avg.Q Tokens    & 112.8 & 211.7 & 102.1 & 107.1 & 376.9   \\
    Avg.A Tokens    & 1 & 2.65 & 1.35 & 22.6 & 73.3    \\
    Avg.AE Tokens    & 232.8 & 313.9 & 211.8 & 241.6 & 372.7    \\
    Avg.No.KC & 2.4 & 2.7 & 2.7 & 2.4 & 2.6     \\
 \bottomrule[1.0pt]
  \end{tabular}
  \caption{Overall statistics of \textbf{CJEval}. S: Subject. Q: Question. KC: Knowledge Concept. AE: Answer Explanation. Here, No.S represents the number of accounts covered under the corresponding question type. Avg.No.KC represents the average number of knowledge concepts associated with each question. Regarding dataset segmentation, the train set, valid set, test set and total set consist of 20,820, 2,106, 3,210 and 26,136 questions, respectively.}
  \label{tab:statistics}
\end{table}

 \begin{figure}[t] 
  \includegraphics[width=\columnwidth]{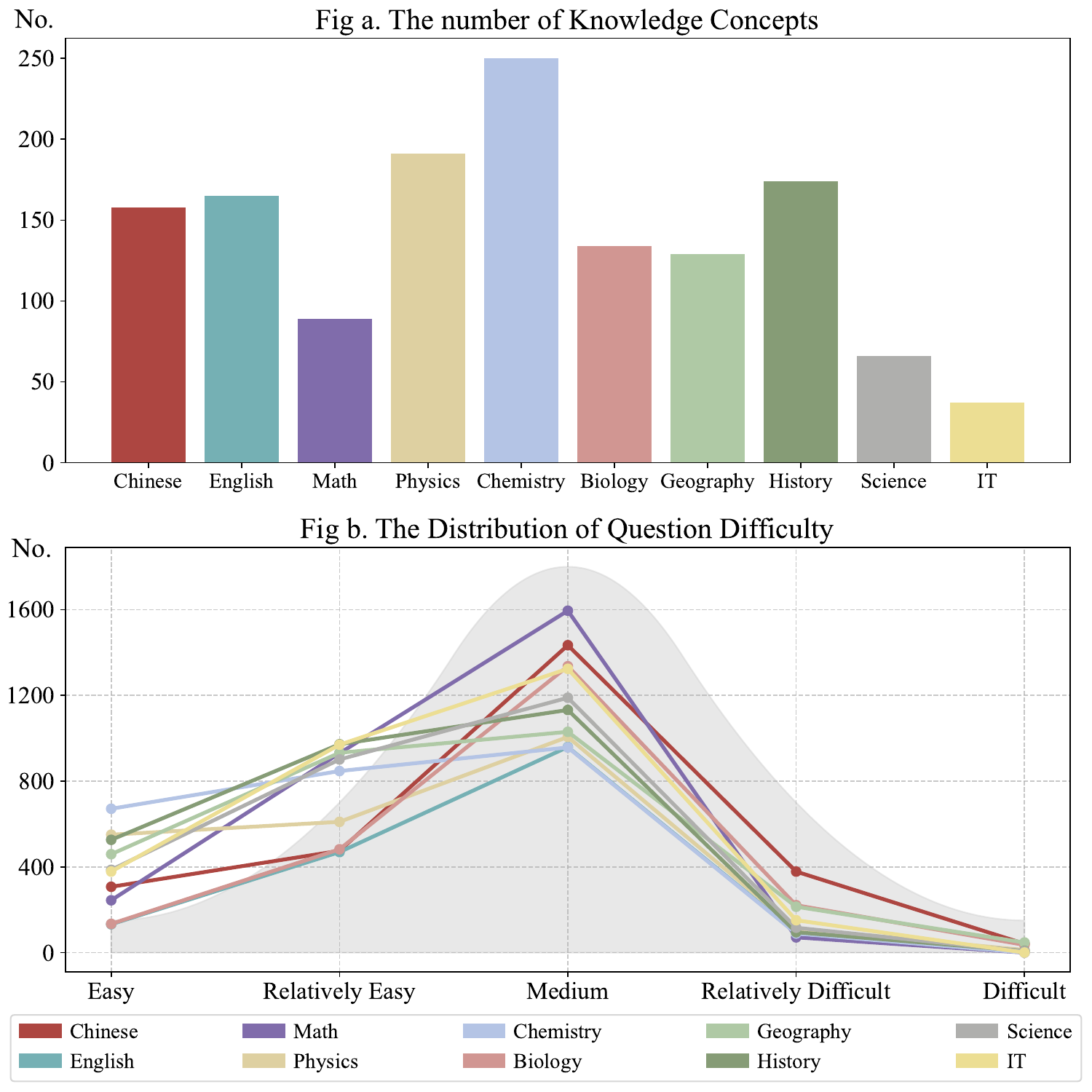}
  \caption{Detailed statistics on question difficulty and knowledge concepts.}
  \label{fig:difficulty}
\end{figure}

The difficulty levels of the questions are classified into five tiers: easy, relatively easy, medium, relatively difficult, and difficult. The difficulty label for each question is annotated based on empirical error rates observed by the online system, which ensures that the difficulty assessment is consistent with the cognitive level of learners (i.e., junior high school students). Each question is also tagged with one or more knowledge concept labels, meticulously derived from educational curriculum guidelines to ensure alignment with pedagogical standards. The annotations pertaining to knowledge concepts and answer explanations have been manually curated with the assistance of pedagogical experts. Note that we used GPT-4 to assist in improving the data to ensure clear representation, and annotators were asked to eliminate potential risks by checking text content.

In Table~\ref{tab:statistics}, we provide the overall statistics of CJEval, broken down by question type. According to the four groups mentioned previously, the dataset comprises 10,270 MCQs (including both SCQs and MRQs), 3,729 TFQs, 6,193 FBQs, and 5,944 AQs. A detailed analysis reveals notable differences in length across these categories. MRQs generally exhibit greater length compared to SCQs, both in terms of the questions themselves and the corresponding answers. For FBQs and AQs, the average question lengths are 107.1 and 376.9, respectively, with average answer lengths of 22.6 and 73.3. This indicates that AQs are the most complex, requiring more extensive responses. To ensure clarity in our dataset, we provide detailed answer explanations that elucidate the reasoning process. 
On average, each question is linked to 2 to 3 knowledge concepts, as detailed in Table~\ref{tab:statistics}. Figure~\ref{fig:difficulty}a illustrates the distribution of these knowledge concepts across various subjects. Additionally, Figure~\ref{fig:difficulty}b shows that the question difficulty distribution approximates a normal curve. This distribution features a higher concentration of questions at intermediate difficulty levels, with fewer questions at the extreme ends of the spectrum.

% https://arxiv.org/pdf/2404.03429
\section{Experiments}
\subsection{Accessed LLMs}
To assess the capabilities of LLMs in educational contexts, our experiments included several open-source models that have been instruction-tuned using Supervised Fine-Tuning (SFT) \citep{ouyang2022training} and Reinforcement Learning from Human Feedback (RLHF) \citep{stiennon2020learning, bai2022training}. Specifically, we utilized the following open-source models: LLaMA3-8B-Instruct\footnote{https://github.com/meta-llama/llama3}, Qwen1.5-Chat series\footnote{https://qwenlm.github.io/blog/qwen1.5}, GLM4-9B-Chat\footnote{https://github.com/THUDM/GLM-4}, Baichuan2-13B-Chat\footnote{https://github.com/baichuan-inc/Baichuan2} and DeepSeek-V2-Chat\footnote{https://github.com/deepseek-ai/DeepSeek-V2}. We also obtained evaluation data through API calls for several proprietary models. These included GPT-3.5-turbo and GPT-4o from OpenAI\footnote{https://openai.com/product}, ERNIE-3.5\footnote{https://cloud.baidu.com}, Baichuan-4\footnote{https://platform.baichuan-ai.com}, Doubao\footnote{https://www.volcengine.com}, Moonshot\footnote{https://www.moonshot.cn} and Hunyuan-pro\footnote{https://cloud.tencent.com/document/product/1729/97731}.

\subsection{Experiment Setups}
\subsubsection{Implementation Details}
{\setlength{\parindent}{0pt}
\textbf{Zero-shot / One-shot Evaluation}
We prioritized using zero-shot settings for three key reasons: First, this approach encourages the models to replicate the problem-solving dynamics typical of real-world exams. Second, certain LLMs are constrained by limited context lengths. Third, the majority of current LLMs have been instruction-tuned, making them adept at adhering to directives and generating outputs in the specified format.
We also compared these with one-shot settings. This approach allows for retrieval-augmented generation, which can introduce similar knowledge to enhance task performance. 
One-shot settings offer an optional strategy for engineering practice, enabling models to better understand and solve subsequent problems by using a single example.

\textbf{Fine-tuning Settings}
We utilized both full-parameter and LoRA-based \citep{hu2021lora} fine-tuning to enhance the models' ability to answer questions and master knowledge across various subjects. The training data include 20,820 samples from our proprietary database, which are sourced similarly to open-source evaluation sets but are non-redundant. The fine-tuning process was conducted on a workstation equipped with 4 NVIDIA Tesla A800 GPUs, each with 80 GB of memory. We utilized the AdamW optimizer, setting the learning rate to \(2 \times 10^{-5}\) for full-parameter training with 1 epoch, and \(3 \times 10^{-4}\) for LoRA training with 5 epochs. We set the warm-up ratio to 0.1 and the weight decay to 0.1. 

}

\begin{table*}[t]
\centering  
\small
\caption{The overall results of the four question-based tasks under zero-shot settings are summarized below. EduScore, a comprehensive metric designed to reflect the performance within the educational system, is displayed at the forefront. }  
\label{tab:Final_result}  
\begin{tabular}{lccccccccccccc}  
\toprule[1.0pt]
\multirow{2}{*}{Model} && \multirow{2}{*}{EduScore} && \multicolumn{3}{c}{KCT} && {QDP} && {QA} && \multicolumn{2}{c}{QG} \\ 
\cline{5-7} \cline{9-9} \cline{11-11} \cline{13-14}
&& && Precision & Recall & F1 && TAcc && Acc && GPT-K & GPT-D \\
\hline
\rowcolor{gray!20} \multicolumn{14}{c}{\textit{Proprietary LLMs}} \\
\hline
GPT-4o (240830) && 75.82 && 60.64 & 79.75 & 64.33 && 82.49 && 67.46 && \textbf{97.34} & \textbf{97.42} \\
% GPT-4 (240409) && 75.30 && 59.13 & 78.34 & 61.23 && 85.96 && 65.91 && \textbf{96.74} & \textbf{98.22} \\
% GPT-3.5 (231106) && 62.68 && 54.50 & 60.57 & 52.15 && 81.54 && 42.87 && 94.31 & 93.64 \\
Moonshot-32k (240508) && 70.17 && 60.14 & 76.68 & 61.48 && 84.35 && 55.66 && 91.95 & 95.46 \\
ERNIE-3.5 (240329) && 71.68 && 68.56 & 72.59 & 64.61 && 80.26 && 59.61 && 92.58 & 96.04 \\
Baichuan-4 (240519) && 71.29 && 58.96 & 80.14 & 62.13 && 83.32 && 59.39 && 90.19 & 94.29 \\
Doubao-pro (240515) && 75.29 && 63.38 & 77.68 & 64.86 &&81.27 && 68.99 && 90.11 & 94.60 \\
Hunyuan-pro (240908) && 76.42 && 63.16 & 77.72 & 65.32 && 88.00 && 67.43 && 93.78 & 94.14 \\
\hline
\rowcolor{gray!20} \multicolumn{14}{c}{\textit{Open-source LLMs}} \\
\hline
LLaMA3-8B && 56.03 && 42.38 & 66.26 & 46.93 && 75.01 && 33.78 && 88.73 & 92.55 \\
Baichuan2-13B && 58.26 && 34.56 & 50.93 & 41.87 && 83.73 && 38.92 && 83.25 & 92.46 \\
GLM4-9B && 70.06 && 55.06 & 75.40 & 56.33 && 82.01 && 62.27 && 83.93 & 90.99 \\
Qwen1.5-14B && 68.49 && 53.12 & 60.12 & 52.48 && 82.12 && 58.37 &&  88.99 & 93.27 \\
Qwen1.5-72B && 73.25 && 61.12 & 73.81 & 60.72 && 85.79 && 64.10 && 89.39 & 93.73 \\
Qwen1.5-110B && 73.10 && 64.74 & 65.38 & 60.67 && 84.32 && 64.84 && 88.85 & 94.60 \\
DeepSeek-V2-236B && 69.51 && 64.80 & 74.53 & 63.46 && 79.90 && 56.31 && 90.88 & 92.30 \\
% \hline
\rowcolor{gray!10} \multicolumn{14}{c}{\textit{Fine-tuned LLMs}} \\
% \hline
Qwen1.5-14B-Full$_{w/AE}$ && 76.54 && 76.16 & 82.17 & 77.77 && 87.73 && 62.43 && 89.74 & 94.93 \\
Qwen1.5-14B-LoRA$_{w/oAE}$ && 78.83 && 79.42 & 78.72 & 77.35 && 87.01 && 69.27 && 90.42 & 92.12 \\
Qwen1.5-14B-LoRA$_{w/AE}$ && \textbf{83.65} && \textbf{83.68} & \textbf{84.54} & \textbf{82.74} && \textbf{88.85} && \textbf{75.92} && 92.92 & 96.73 \\

\bottomrule[1.0pt]
\end{tabular}  
\end{table*}

\subsubsection{Evaluation Metrics}
Considering that different tools are implemented by combining various tasks in real systems, our evaluation metrics for the four question-based tasks are as follows:

For KCT, we calculated precision, recall, and F1-score for each subject. By default, we use a text similarity model to retrieve the top 20 most similar knowledge concepts, denoted as $K_{q_i}^{\text{top20}}$, from the set $K$. This step ensures that the ground truth $K_{q_i}$ is included in $K_{q_i}^{\text{top20}}$ through manual adjustment, thus focusing on the impact of downstream evaluation on the models' capabilities.
  
For QDP, we use "Tolerant Accuracy" (TAcc) as the evaluation metric instead of Mean Square Error (MSE), because small errors within one level are not perceptually significant to the users. Given the predicted difficulty level \(\hat{dif}_i\) and the true difficulty level \(dif_i\), we define a tolerance function \(\delta(\hat{dif}_i, dif_i)\) as follows:
\[
\delta(\hat{dif}_i, dif_i) = 
\begin{cases} 
1 & \text{if } |\hat{dif}_i - dif_i| \leq 1 \\
0 & \text{otherwise}
\end{cases}
\]
The TAcc is then defined as:
\[
\text{TAcc} = \frac{1}{N} \sum_{i=1}^{N} \delta(\hat{dif}_i, dif_i)
\]
where \(N\) is the total number of samples.

For QA, different question types were evaluated using different methods. MCQs and TFQs employed regular expression matching, with a full match considered correct. Inspired by Chateval \citep{chan2023chateval}, GPT-4 was used as a reviewer for FBQs and AQs to determine correctness.

For QG, we engaged GPT-4o to evaluate the quality of the generated questions from two aspects: whether the question content included all required knowledge concepts and whether the difficulty level met the requirements. We calculated the proportion of questions meeting these criteria as evaluations, referred to as GPT-K and GPT-D.

Subsequently, we define \textbf{EduScore} as the comprehensive score across the four tasks:
\begin{equation*}
\scriptsize
\begin{aligned}
\text{\large EduScore} = & \ 0.4 \times \text{ACC}_{\text{QA}} + \\
& 0.2 \times \left( \text{F1}_{\text{KCT}} + \text{TACC}_{\text{QDP}} + \frac{\text{GPT-K}_{\text{QG}} + \text{GPT-D}_{\text{QG}}}{2} \right)
\end{aligned}
\end{equation*}
Note that QA tasks account for 40\% of the total score due to their critical significance. This emphasis is placed because QA tasks rigorously assess the system's capability to comprehend and generate precise responses to user inquiries, which are fundamental to the system's overall efficacy.

\subsection{Results and Analysis}
To identify the strengths and weaknesses of different LLMs, we present and analyze their performance from multiple perspectives.

\subsubsection{Multi-task Evaluation}
Table~\ref{tab:Final_result} presents the results of four question-based tasks across different LLMs. The results indicate that Hunyuan-pro, GPT-4o, and Doubao-pro are ahead of other proprietary and open-source models in terms of EduScore. Notably, the fine-tuning of Qwen1.5-14B significantly improved its performance in various tasks, with the highest EduScore being obtained by Qwen1.5-14B-LoRA$_{w/AE}$. This phenomenon aligns with the understanding that high-quality data could help LLMs develop capabilities effectively.

\begin{figure*}[t]
  \includegraphics[width=2\columnwidth]{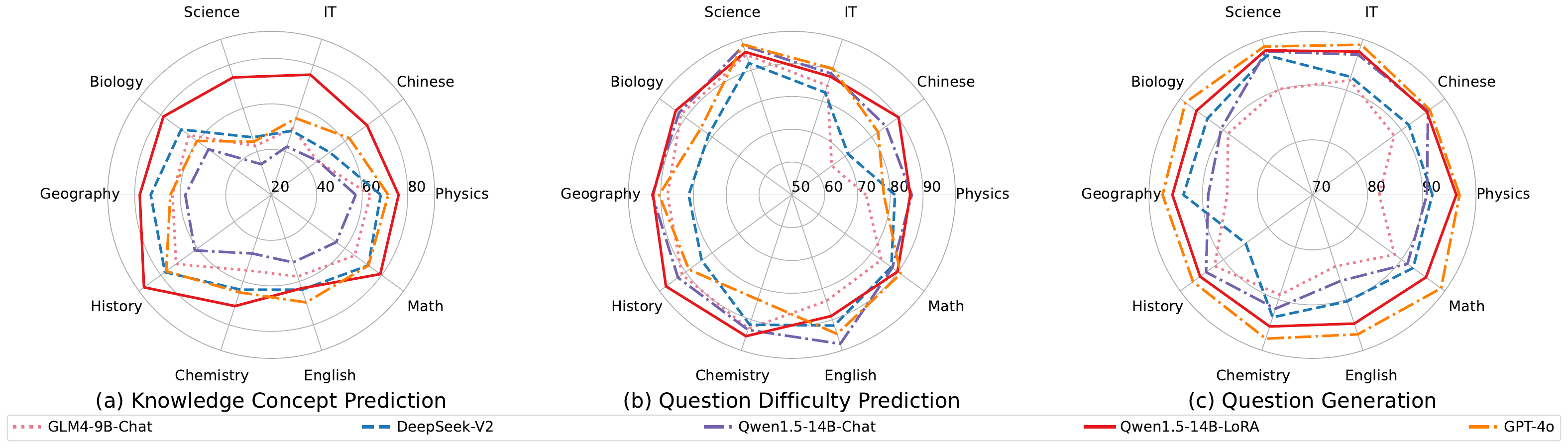}
  \caption{The performance of models in KCT, QDP, and QG tasks across ten subjects.}
  \label{fig:subject}
\end{figure*}

\begin{figure*}[t]
  \includegraphics[width=2\columnwidth]{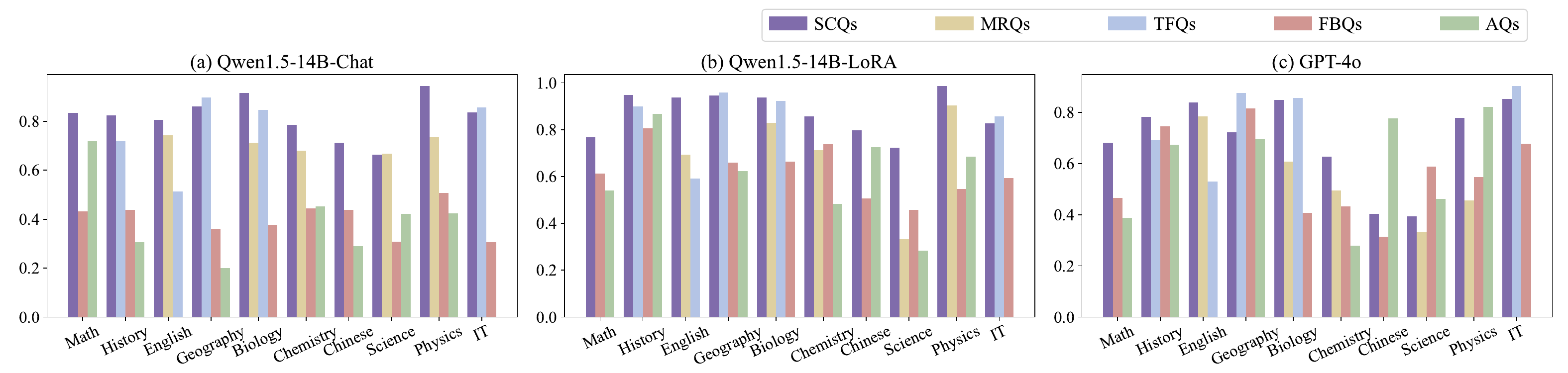}
  \caption{The performance of models in different question types across ten subjects.}
  \label{fig:qa}
\end{figure*}

Specifically, QA and KCT are particularly challenging tasks, with models that were not fine-tuned generally scoring below 70. In QG, these models demonstrated a strong ability to mimic the style and content of existing questions, achieving scores around 90 in both GPT-K and GPT-D. Among the fine-tuned LLMs, the model trained with answer explanations (AE) consistently outperformed the one without. This improvement can be attributed to the inclusion of reasoning processes in the answer explanations, which help the model better understand the underlying logic and context of the questions. Interestingly, full-parameter fine-tuning did not yield the best results. This could be due to several factors. One possibility is that full-parameter tuning may lead to overfitting, where the model becomes too specialized in the training data and loses its generalization capability. Another reason could be the increased computational complexity and resource requirements associated with full-parameter tuning, which might not always translate to proportional performance gains. 
% Instead, more targeted approaches like LoRA-based fine-tuning, especially with answer explanations, seem to strike a better balance between performance and resource efficiency.
In summary, we prefer to fine-tune smaller models on high-quality data rather than simply relying on larger language models. When it comes to training, low-loss, high-efficiency schemes like LoRA are highly valuable.

\subsubsection{Multi-subject Evaluation}
To more precisely demonstrate the models' capabilities across various subjects, Figure~\ref{fig:subject} selects several top models and lists their detailed performance metrics across various subjects. Using GPT-4o as a reference baseline,  it is clear that models such as Qwen, which specialize in Chinese, demonstrate strong capabilities in educational tasks.
All models demonstrate robust performance in subjects such as history, geography, and biology, which primarily rely on memory skills. Conversely, performance drops significantly in subjects that demand higher-order reasoning skills, such as math, physics, and chemistry. This decline can be attributed to the scarcity of existing data and the need for extensive inference and computation. Additionally, anomalies in predicting scientific knowledge concepts may arise from their customized definitions and the models' inability to form associations through generalized understanding. This underscores the necessity for further training of LLMs in applied contexts.

Table~\ref{tab:Final_result} demonstrates that, even after fine-tuning, the accuracy metrics for QA tasks remain below 76\%. To further analyze this, we have detailed the performance across different question types in various subjects in Figure~\ref{fig:qa}. MRQs are more challenging than SCQs due to the necessity for complex logical reasoning to evaluate multiple options. TFQs are relatively easier for the models to comprehend. The accuracy for AQs remains low, indicating that these models struggle with tasks requiring the generation of precise and contextually appropriate responses. This finding underscores the need to enhance reasoning and language generation capabilities. Notably, GPT-4o performs well on AQ tasks, which can be attributed to its sophisticated architecture and extensive pre-training.

Overall, Chinese models exhibit an advantage in Chinese language tasks, while GPT-4o excels in English language tasks. The fine-tuned Qwen-14B model has become comparable to GPT-4o, even surpassing it in several subjects. This phenomenon underscores the feasibility of deploying large language models in various commercial applications.

\subsubsection{Zero-shot vs. One-shot}

\begin{table}[H]
\setlength\tabcolsep{2pt}
\centering  
\small
\begin{tabular}{lcccccccc}  

\hline  
\multirow{2}{*}{Model} && \multirow{2}{*}{shot} && \multicolumn{3}{c}{KCT} && {QDP} \\ \cline{5-7} \cline{9-9}
&& && P & R & F1 && TAcc  \\  
\hline
\multirow{3}{*}{\small GPT-4o} && zero && 60.64 & 79.75 & 64.33 && \textbf{82.49}   \\
&& one && \textbf{63.45} & \textbf{83.64} & \textbf{69.09} && 80.43   \\
 \hline
 \multirow{3}{*}{\small Hunyuan-pro} && zero && 63.16 & 77.72 & 65.32 && \textbf{88.00}   \\
 && one && \textbf{65.40} & \textbf{80.66} & \textbf{68.79} && 85.96   \\
 \hline
 \multirow{3}{*}{\small Baichuan4} && zero && 58.96 & 80.14 & 62.13 && \textbf{83.32}   \\
 && one && \textbf{61.84} & \textbf{82.47} & \textbf{66.01} && 81.07   \\
 \hline
 \multirow{3}{*}{\small Doubao-pro} && zero && 63.38 & 77.68 & 64.86 && \textbf{79.27}   \\
 && one && \textbf{64.77} & \textbf{78.41} & \textbf{64.92} && 76.24   \\
  \hline
 % && few && 36.95 & 54.28 & 40.18 && 43.52   \\
 \multirow{3}{*}{\small DeepSeek-V2} && zero && 64.80 & 74.53 & 63.46 && \textbf{79.90}   \\
 && one && \textbf{64.80} & \textbf{74.84} & \textbf{64.83} && 78.64   \\
  \hline
\multirow{3}{*}{\small LLaMA3-8B} && zero && 42.38 & \textbf{66.26} & \textbf{46.93} && \textbf{75.01}   \\
 && one && \textbf{45.86} & 62.01 & 43.27 && 66.89   \\
 \hline
\multirow{3}{*}{\small Qwen1.5-14B} && zero && 53.12 & \textbf{60.12} & 52.48 && \textbf{82.12}   \\
 && one && \textbf{54.53} & 58.80 & \textbf{53.59} && 81.38   \\
 % && few && \textbf{53.63} & 64.17 & \textbf{53.84} && 72.96  \\
\hline
\multirow{3}{*}{\small Qwen1.5-14B-LoRA} && zero && 81.77 & 82.11 & 80.38 && \textbf{88.85}   \\
 && one && \textbf{83.21} & \textbf{82.90} & \textbf{81.02} && 84.38   \\
 % && few && \textbf{86.27} & \textbf{84.87} & \textbf{85.56} && 76.23   \\

\hline
% \multirow{3}{*}{\small GPT-3.5} && zero && 54.50 & 60.57 & 52.15 && \textbf{76.24}   \\
%  && one && \textbf{56.55} & \textbf{61.98} & \textbf{59.14} && 77.93   \\
%  % && few && 53.52 & \textbf{62.47} & 57.64 && 1   \\
% \hline

\end{tabular}  
\caption{Results on different prompting strategies based on different LLMs.} 
\label{tab:Shot results} 
\end{table}

% From the experimental results shown in \textbf{Table \ref{tab:Shot results}}, we can observe that although the improvements in some individual subjects were weak or even slightly decreased, overall, using a small number of examples for prompting provided effective additional information for each task, thus improving the model's performance. The reason might be that the example samples helped the model better understand the task context, provided a comparative object for difficulty evaluation, guided the solution approach, and allowed the model to generate higher-quality questions through imitation and modification. Thus, using in-context demonstrations indeed provides additional advantages.
To empirically study the impact of one-shot demonstrations, we selected one-shot examples for KCT from the development dataset. These examples were chosen based on having the same subject as the target question and at least one overlapping knowledge point within the $K_{q_i}^{\text{top20}}$. For QDP, the one-shot examples were selected based on having the same question type and subject, with a medium difficulty level as the standard.
From the experimental results shown in Table~\ref{tab:Shot results}, we can observe that using a small number of examples to provide effective additional information for KCT tasks can improve the performance of the model in most cases. However, in QDP, performance declined. This decline might be attributed to the fact that the one-shot examples did not adequately capture the complexity and variability of question difficulty, leading to less effective guidance for the model. Based on the above, the RAG mentioned in the system design, while important in real production, should be applied judiciously.

\section{Related Work}
{\setlength{\parindent}{0pt}
\textbf{Intelligent tutoring systems (ITS)}
Intelligent Tutoring Systems (ITS) have been a focal point of research in educational technology for several decades, aiming to provide personalized instruction and feedback akin to that offered by a human tutor \citep{nwana1990intelligent, muirhead2000online}. These systems are particularly pivotal in the STEM field, where enhancing educational accessibility through online platforms has proven to be a cost-effective approach \citep{chirikov2020online}. Moreover, problem-centered instruction has been identified as a significant trend in current research, emphasizing the importance of addressing real-world issues within educational contexts \citep{guo2021evolution}.
Research on ITS is inherently multidisciplinary, necessitating that researchers adopt innovative ideas and maintain a high level of perseverance in the face of challenging research tasks \citep{shneider2009four}. This comprehensive approach helps in creating systems that not only mimic the pedagogical guidance of human tutors but also enhance the educational experience through personalized learning trajectories.

\textbf{Advancements in Large Language Models}
Large Language Models (LLMs) have made remarkable strides in recent years, showcasing their ability to generate human-like text, answer complex questions, and perform a wide array of natural language processing (NLP) tasks. Models such as GPT \citep{achiam2023gpt}, LLaMA \citep{touvron2023llama}, Baichuan \citep{yang2023baichuan}, Qwen \citep{bai2023qwen}, GLM \citep{du2022glm}, and DeepSeek-V2 \citep{deepseekv2} have set new standards, demonstrating exceptional performance across various benchmarks. These models have been widely adopted in both academic and industrial settings, highlighting their versatility and potential. However, the evaluation of LLMs is a multifaceted challenge that extends beyond mere accuracy in answering questions. It involves assessing their ability to understand context, generate coherent and relevant responses, and apply specialized knowledge effectively.

\textbf{Evaluation Benchmarks} 
The evaluation of LLMs has traditionally relied on established NLP benchmark datasets, such as SQuAD \citep{rajpurkar2016squad}, GLUE \citep{wang2018glue}, and SuperGLUE \citep{wang2019superglue}, which focus on specific tasks like question answering, sentiment analysis, and text classification. Recent efforts have focused on utilizing human-designed exams to evaluate LLMs, with datasets like MMLU \citep{hendrycks2020measuring}. There are also datasets in Chinese that focus on expanding question types, such as GAOKAO \citep{zhang2023evaluating}, CGEval \citep{zeng4578709evaluating}, and LHMKE \citep{liu2024lhmke}. 
% GAOKAO includes authentic Chinese college entrance examination questions, consisting of 1,781 objective questions and 1,030 subjective questions. In contrast, CG-Eval evaluates Chinese text generation capabilities through 11,000 subjective questions spanning three distinct categories. Multidimensional datasets are still scarce, ensuring that evaluations are not limited to a narrow set of tasks, thus providing a more comprehensive view of model capabilities.
}
% In summary, 

% While significant progress has been made in the development and evaluation of LLMs, there remains a need for more comprehensive and nuanced evaluation frameworks that can better capture the complexities of real-world educational scenarios. This work aims to address these gaps by proposing a next-generation online education platform framework and a multidimensional dataset that can provide a more rigorous assessment of LLM capabilities in educational settings.

\section{Conclusion}
In this work, we developed \textbf{CJEval}, a dataset sourced from Chinese Junior High School Examinations, which features diverse exam questions with detailed annotations. CJEval ensures reliability, validity, and comprehensiveness by encompassing four application-level tasks across ten subjects. This dataset provides a robust basis for assessing educational competencies and demonstrates the practicality of employing LLMs within educational contexts. We anticipate that CJEval will contribute significantly to future advancements of LLMs, particularly in handling educational tasks.

\section{Limitation}
Online education systems encompass a diverse array of applications, such as personalized recommendations and adaptive learning systems, which are beyond the scope of this paper. In addition, our study concentrates on the capabilities of LLMs in handling question-based tasks. We did not delve into the advantages and disadvantages of RAG methods. Future research should explore the integration of RAG methods to fully understand their impact on the performance and applicability of LLMs in instructional applications.

\section{Ethics}
CJEval is derived from actual junior high school test questions, which have been meticulously rewritten and scrutinized. The CJEval dataset is intended solely for academic and research purposes. Commercial use or any misuse that deviates from these intended purposes is strictly prohibited. Adhering to these guidelines is essential to maintain the dataset's integrity and ensure ethical use.

\bibliography{custom}

% \appendix

% \onecolumn
% \section{Prompt} \label{sec:prompt}
% Figure~\ref{fig:prompt} shows the details of our prompt settings. For the convenience of reading, we have used equivalent English expressions for display.
% % \begin{figure*}[\linewidth]
% %   \includegraphics[width=0.98\linewidth, height=1.15\linewidth]{source/Prompt.pdf}
% %   \caption{Inference and Evaluation Prompts for Four Tasks}
% %   \label{fig:prompt}
% % \end{figure*}
% % \section{Example Appendix}
% % \label{sec:appendix}

% % This is an appendix.

\end{document}